\documentclass[letterpaper, 10 pt, journal]{ieeetran}

\IEEEoverridecommandlockouts                              

\usepackage{color}
\usepackage{soul}

\title{
DAPPER: Discriminability-Aware Policy-to-Policy Preference-Based Reinforcement Learning for Query-Efficient Robot Skill Acquisition
}

\author{Yuki Kadokawa$^{1}$, Jonas Frey$^{2}$,  Takahiro Miki$^{2}$, Takamitsu Matsubara$^{1}$, and Marco Hutter$^{2}$
\thanks{
    This work was supported by JSPS KAKENHI Grant Numbers JP23KJ1585 and JP24K03018.
    \textit{(Corresponding author: Yuki Kadokawa.)}
    $\ $ \textsuperscript{1} Yuki Kadokawa and Takamitsu Matsubara are with Nara Institute of Science and Technology, Nara 630-0192, Japan, kadokawa.yuki@naist.ac.jp, takam-m@is.naist.jp. 
    \textsuperscript{2} Jonas Frey, Takahiro Miki, and Marco Hutter are with ETH Zurich, Zurich 8092, Switzerland, jonfrey@ethz.ch, tamiki@ethz.ch, mahutter@ethz.ch.
}%
}

\usepackage{setspace}
\usepackage{comment}
\usepackage{subfigure}
\usepackage{booktabs}
\usepackage{graphicx}
\usepackage{amsmath}
\usepackage{amsfonts}
\usepackage{algorithmic}
\usepackage[font=footnotesize]{caption}
\usepackage{siunitx}
\usepackage{bm}
\usepackage[space, compress]{cite}
\usepackage[ruled,vlined]{algorithm2e}
\usepackage{algorithm2e,setspace}
\usepackage{textcomp}
\usepackage{mathcomp}
\usepackage{tabularx}
\usepackage{here}
\usepackage[colorlinks=true, linkcolor=blue, citecolor=blue, urlcolor=blue]{hyperref}

\newcommand{\tabref}[1]{\hyperref[#1]{Table~\ref*{#1}}}
\newcommand{\equref}[1]{\hyperref[#1]{Eq.~(\ref*{#1})}}
\newcommand{\figref}[1]{\hyperref[#1]{Fig.~\ref*{#1}}}
\newcommand{\chapref}[1]{\hyperref[#1]{Section~\ref*{#1}}}
\newcommand{\algref}[1]{\hyperref[#1]{Algorithm~\ref*{#1}}}
\newcommand{\apperef}[1]{\hyperref[#1]{Appendix~\ref*{#1}}}

\begin{document}


\maketitle

\begin{abstract}
    Preference-based Reinforcement Learning (PbRL) enables policy learning through simple queries comparing trajectories from a single policy. While human responses to these queries make it possible to learn policies aligned with human preferences, PbRL suffers from low query efficiency, as policy bias limits trajectory diversity and reduces the number of discriminable queries available for learning preferences.
    This paper identifies {\it{preference discriminability}}, which quantifies how easily a human can judge which trajectory is closer to their ideal behavior, as a key metric for improving query efficiency.
    To address this, we move beyond comparisons within a single policy and instead generate queries by comparing trajectories from multiple policies, as training them from scratch promotes diversity without policy bias.
    We propose Discriminability-Aware Policy-to-Policy Preference-Based Efficient Reinforcement Learning (DAPPER), which integrates preference discriminability with trajectory diversification achieved by multiple policies.
    DAPPER trains new policies from scratch after each reward update and employs a discriminator that learns to estimate preference discriminability, enabling the prioritized sampling of more discriminable queries.
    During training, it jointly maximizes the preference reward and preference discriminability score, encouraging the discovery of highly rewarding and easily distinguishable policies.
    Experiments in simulated and real-world legged robot environments demonstrate that DAPPER outperforms previous methods in query efficiency, particularly under challenging preference discriminability conditions.
    A supplementary video that facilitates understanding of the proposed framework and its experimental results is available at: \url{https://youtu.be/lRwX8FNN8n4}
\end{abstract}


    
 \section{Introduction}

    \IEEEPARstart{L}{egged} locomotion has received considerable attention, with growing interest in enabling robots to acquire walking behaviors that reflect human preferences rather than relying solely on task-specific objectives \cite{legged_gym, IL_wasabi}. Existing approaches mainly rely on reinforcement learning \cite{ConstraintPPO_1,ConstraintPPO_2} or imitation learning \cite{IL, IL_wasabi}. Reinforcement learning requires task-specific reward parameter tuning that depends heavily on expert knowledge, while imitation learning relies on demonstrations that demand operator proficiency with robot interfaces and motion strategies. Both introduce substantial costs for users, making it difficult to realize locomotion behaviors aligned with individual preferences. In this paper, we focus on quadruped locomotion and propose a preference-based learning framework that reduces user effort while shaping policies according to human preferences.

\begin{figure}[t]
    \centering
    \includegraphics[width=0.95\columnwidth]{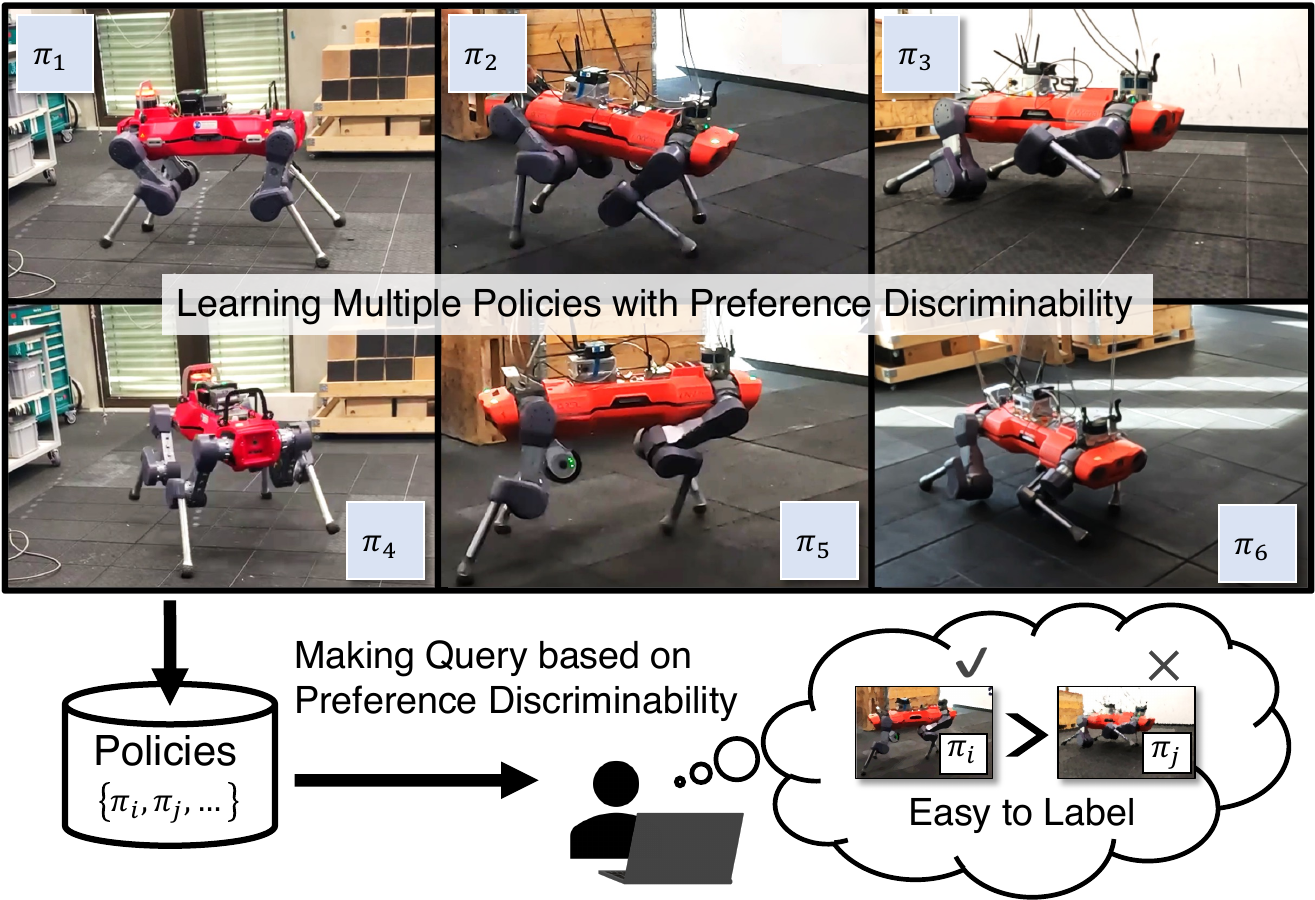}
    \caption{
        Overview of the proposed learning framework:
        Multiple policies with diverse feature representations are sequentially learned. Queries are generated by sampling from these policies based on human preference discriminability, ensuring easier labeling. This approach enhances query efficiency by increasing the number of discriminable queries.
    }
    \label{fig:promotional_photo}
\end{figure}

    Preference-based Reinforcement Learning (PbRL) has gained attention as it enables the learning of control policies through simple query responses \cite{pbrl,PbRL_auto_label}. In typical PbRL frameworks, two state-action trajectories are sampled from interactions between a single policy and the environment, and a human selects the more preferred one. Based on these preference labels, a reward model that approximates human preferences is learned, and the control policy is updated through reinforcement learning. By iteratively updating the reward model with new query responses and optimizing the policy accordingly, PbRL gradually learns a policy that reflects human preferences \cite{pbrl_noisyPreference,pbrl_misalignment}.

    A key challenge in PbRL is that diverse state-action trajectories are required to efficiently learn human preferences via queries \cite{pbrl_problems, pbrl_survey}. Ideally, each update of the reward model should lead to the acquisition of new policies that can generate novel state-action trajectories. However, when using a single policy, policy bias \cite{policy_bias1,policy_bias2} becomes a critical problem. 
    Policy bias refers to the limited behavioral diversity that arises when learning from a single policy, as new policies are trained as incremental updates of previous ones. Because each policy is a continuation of the last, their behaviors tend to be strongly correlated, making it difficult to explore new actions and fully exploit updated reward functions.
    As a result, the collected trajectories tend to be similar, and behavioral diversity is significantly limited.

    To address this issue, we focus on the ease of labeling and introduce {\it preference discriminability} as a key metric. 
    This metric quantifies whether a human can determine how much two trajectories in a query are closer to their ideal behavior.   
    When one trajectory is clearly closer to the ideal, labeling is straightforward. 
    However, when both trajectories are similarly undesirable or equally good, humans may struggle to provide a clear preference, not because the trajectories themselves are indistinguishable, but because their underlying preferences are indistinguishable.
    By generating queries with high preference discriminability, we present clearer differences between trajectories, making them easier to label and thereby improving query efficiency, as a greater number of discriminable queries can be used to learn the human preference.
    
    Building on this perspective, we focus on moving beyond comparisons within a single policy and instead generate queries by comparing trajectories from different policies as shown in \figref{fig:promotional_photo}. Training multiple policies encourages the emergence of more diverse behaviors, since learning different policies from scratch is not constrained by the policy bias. 
    This approach avoids the tendency of gradual policy updates to produce similar behaviors and supports the generation of richer and more varied state-action trajectories. 
    However, simply increasing diversity does not guarantee that the resulting queries will be easy for humans to label. To ensure that diversity leads to meaningful and discriminable comparisons, we further introduce a preference discriminability-aware query generation mechanism that estimates the discriminability of trajectory pairs.

    To this end, we propose a novel framework, Discriminability-Aware Policy-to-Policy Preference-Based Efficient Reinforcement Learning (DAPPER), which integrates two key ideas: preference discriminability and trajectory diversification achieved by multiple policies. 
    DAPPER approximates preference discriminability as a learnable function by training a discriminator that estimates whether a given query is likely to be distinguishable by humans, based on the history of queries labeled as indistinguishable.
    During training, DAPPER probabilistically samples queries based on their predicted preference discriminability, promoting the selection of discriminable queries.
    To enhance trajectory diversity, DAPPER trains multiple policies. Each time the reward model is updated, a new policy is retrained from scratch. 
    Finally, during policy learning, we not only maximize for the reward of human preference but also incorporate the preference discriminability score as an auxiliary reward. This dual-objective approach encourages the discovery of policies that yield both high-reward and easily distinguishable behaviors.
    
    We demonstrated the effectiveness of DAPPER in a legged robot simulator and real-world robot environment. DAPPER achieves successful policy learning with fewer queries than previous methods. Furthermore, compared to previous methods, only DAPPER consistently learn policies under the conditions that preference discriminability becomes increasingly challenging. We also applied the learned policies to real-world environments.

\section{Related Work}
    \label{s:related_works}

    \subsection{Learning Control Policies}

        This section describes previous methods for learning control policies and the differences from our approach.

        \textbf{Imitation Learning:} 
            Learning from human demonstrations has been explored in previous studies \cite{IL, IL_wasabi}. However, creating demonstrations for complex mechanisms such as quadrupedal robots is challenging. While recent efforts have focused on improving demonstration efficiency by refining demonstration methods, these approaches still require demonstrators to have prior knowledge of the policy features being taught. Moreover, effective data collection depends on a clear understanding of the desired policy features beforehand \cite{IL_wasabi}. For non-experts, predicting and defining the precise behavior of a quadrupedal robot in advance is difficult, making it challenging to acquire the intended policy features.

        \textbf{Reinforcement Learning:}
            Reinforcement Learning (RL) enables robots to autonomously acquire arbitrary policies through trial-and-error interactions with the environment by designing a reward function \cite{ConstraintPPO_1,ConstraintPPO_2}.
            However, designing an appropriate reward function that results in the desired behavior requires a carefully designed reward, often incorporating multiple auxiliary terms and considerable manual effort.
            Moreover, reward function design requires expert knowledge, making it difficult for non-experts to define rewards that effectively capture their intended requirements. 
            As a result, acquiring policies resulting in specific motion patterns through RL remains difficult for non-experts.

        \textbf{Preference-based Reinforcement Learning:}
            Potentially, even non-experts can evaluate and distinguish desirable robot behaviors by responding to video-based comparison queries, enabling the learning of policies that align with human preferences \cite{pbrl,pbrl_survey}. However, PbRL requires collecting a large number of queries, making policy learning infeasible within a practical timeframe. To the best of our knowledge, this paper is the first to apply PbRL to learning quadrupedal robot locomotion. To achieve this, we aim to improve query efficiency, enabling policy learning within a reasonable timeframe.

    \subsection{Improving Query Efficiency in PbRL}
        This section reviews previous works for improving query efficiency in PbRL and discusses the differences from our work.

        \textbf{Data Augmentation:}
            In this approach, in addition to human annotations, the preference reward model under training is also used to automatically assign labels. Specifically, when a query sample is fed into the model and the predicted probability (ranging from 0 to 1) that one trajectory is preferred over the other shows a clear margin, the model's prediction is used as the label instead of querying a human. This enables automatic dataset expansion and reduces the need for human responses, thereby improving query efficiency \cite{PbRL_auto_label, pbrl_survey}.

        \textbf{Robustifying to Incorrect Human Feedback:}
            Incorrect human feedback can mislead the learning process, reducing the effectiveness of each query. To improve query efficiency under such conditions, robust learning techniques are employed to reduce sensitivity to label noise. These include filtering out inconsistent responses and using robust loss functions that reduce the influence of potentially incorrect labels. By mitigating the negative impact of noisy data, these methods enable more reliable policy updates from fewer queries, thereby enhancing overall query efficiency \cite{pbrl_noisyPreference,pbrl_misalignment}.

        \textbf{Uncertainty Sampling:}
            Uncertainty sampling aims to improve query efficiency by prioritizing queries near the decision boundary of the preference reward model, where the output probability is close to $0.5$. These queries are expected to provide the most informative feedback for refining the reward function. By focusing on ambiguous cases, the model can collect targeted training data that reduces uncertainty and sharpens its decision boundaries \cite{pbrl_problems,pbrl_survey}.

        \textbf{Active Reward Learning:}
            This approach collects queries from a pre-generated trajectory dataset and trains a policy only once using the reward function learned from those queries. 
            Following information-theoretic objective, it implicitly favors trajectory pairs that are easier for humans to distinguish, since indistinguishable queries yield little information gain.
            However, because it depends entirely on a fixed dataset of candidate trajectories, it assumes that optimal and easily comparable pairs already exist in the dataset. Without a mechanism to generate new trajectories through policy optimization, it cannot expand behavioral variations for more discriminable queries \cite{mutual_info_query_selection, pbrl_problems}.

        \textbf{Reward Re-calculation:}
            This approach improves query efficiency by recalculating the reward values of previously collected trajectories, enabling the reuse of past dataset without requiring additional queries. The agent can effectively expand its training dataset and refine the reward model more rapidly. This helps reduce the reliance on new annotations and contributes to more stable policy learning \cite{pebble,pbrl_problems}.

        \textbf{Ours:}
            While previous approaches improve query efficiency by leveraging model confidence \cite{PbRL_auto_label}, addressing annotation noise \cite{pbrl_problems}, actively selecting informative queries \cite{pbrl_survey}, selecting queries based on information gain \cite{mutual_info_query_selection}, or reusing past data through reward re-calculation \cite{pebble}, they do not explicitly account for the labeling difficulty experienced by human annotators.
            In contrast, our framework improves query efficiency by modeling and leveraging human preference discriminability, which helps reduce indistinguishable queries and enhances query efficiency.
            This is particularly important in tasks where distinguishing task features is inherently difficult. If a query is not distinguishable to the human, a preference label cannot be obtained.

\section{Preliminaries}

    \subsection{Reinforcement Learning}
        
        Reinforcement Learning (RL) assumes a scenario in which an agent takes actions through observed state \cite{ConstraintPPO_1,ConstraintPPO_2}. The agent receives feedback about its actions in the form of a reward. The goal of the agent is to learn actions to maximize total reward. 
        The standard formalization of a RL problem builds on the notion of a Markov Decision Process (MDP) and consists of: A continuous state space $S\subseteq\mathbb{R}^n$ which represents an observation from the environment; A continuous action space $A\subseteq\mathbb{R}^m$ representing the set of actions available to the agent; state transition probability $\mathcal{P}:S\times A\rightarrow P(S)$, where $P(S)$ denotes the probability that action $a$ in state $s$ leads the agent to state $s'$; $\gamma \in [0,1)$ is the discount factor.
        The agent interacts with the environment by taking an action $a\in A$ based on the current state $s\in S$, receiving a reward $r(s,a)$, and transitioning to a new state $s'\in S$ according to the probability distribution $\mathcal{P}(\cdot\mid s,a)$.
        
        The agent moves through the state space by repeatedly taking actions, thereby generating a trajectory $\tau=(s_0,a_0,s_1,\dots,s_t,a_t,s_{t+1})$, where $s_0$ is the state the agent starts with, $a_t$ is the action it takes in state $s_t$, and $s_{t+1}$ is the next state produced by this action. 
        
        The most common task in reinforcement learning is to learn a policy $\pi:S\rightarrow A$ that prescribes the optimal action for the agent in each state. More specifically, the objective is often defined as maximizing the total rewards: $V_\pi(s)=\mathbb{E}_{\pi,\mathcal{P}}\Bigl[\textstyle \sum_{t=0}^{\infty}\gamma^t\, r(s_t,a_t) \,\Bigl|\, s_0=s\Bigr]$.
        Accordingly, common approaches have focused on learning an optimal policy by maximizing the value function $V$. Among various policy optimization algorithms, Proximal Policy Optimization (PPO) is recognized as an effective means to achieve this objective \cite{ConstraintPPO_1,ConstraintPPO_2}.

\begin{figure}[t]
    \centering
    \includegraphics[width=0.99\columnwidth]{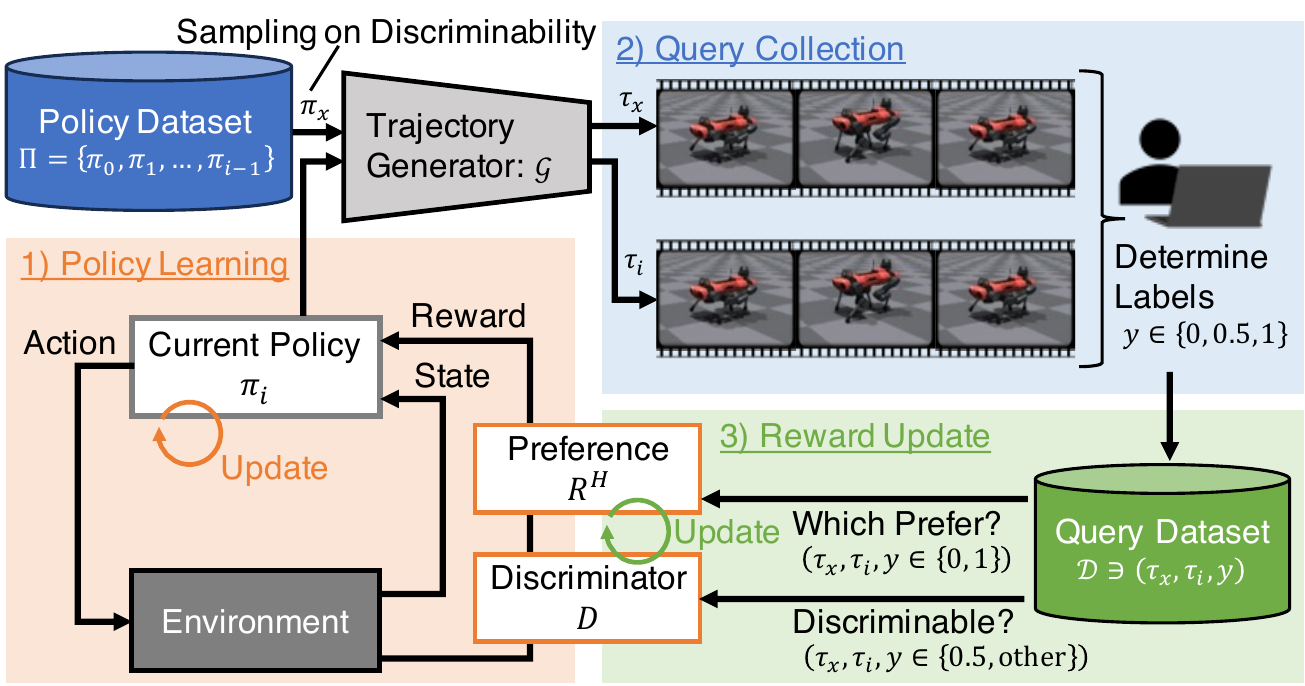}
    \caption{
        Learning process of DAPPER:
        Our framework consists of three steps.
        (1) Learning current policy $\pi_i$ with reward $r$.
        (2) Query collection. This step makes queries from current policy $\pi_i$ and sampled policy from policy dataset $\Pi$ based on discriminability. Then, trajectories of each policy are generated $\tau_x$ and $\tau_i$, and queries are labeled $y$.
        (3) Reward Update. This step trains the preference reward model $R^H$ and discriminator $D$ from query dataset $\mathcal{D}$.
    }
    \label{fig:proposed_method}
\end{figure}

    \subsection{Preference-based Reinforcement Learning}
        \label{chap:pbrl}
        In Preference-based Reinforcement Learning (PbRL), the agent learns a reward function based on human feedback and then uses this reward function to learn the optimal policies \cite{pbrl}. The process is as follows.

        First, the agent collects state-action trajectories $\tau$. This transition is provided as preferences over pairs of trajectories, expressed as $P(\tau_i \succ \tau_j)$, where $P(\tau_i \succ \tau_j)$ represents the probability that trajectory $\tau_i$ is preferred over trajectory $\tau_j$. Basically, for each trajectory of a query pair, PbRL utilizes an interval clipped from the trajectory to a specific step length.
        
        Human preference is modeled by logistic regression as follows:
        \begin{eqnarray}
            \label{eq:previous_preference}
            P(\tau_i \succ \tau_j) = \frac{\exp(r(\tau_i))}{\exp(r(\tau_i)) + \exp(r(\tau_j))}
        \end{eqnarray}
        where $r(\tau)$ is the cumulative reward function for trajectory $\tau$.
        
        Given the feedback data $\{(\tau_i, \tau_j)\}_{k=1}^N$, the reward function $r(\tau)$ is optimized by minimizing the binary cross-entropy loss function:
        \begin{eqnarray}
            \label{eq:previous_loss}
            \mathcal{L} &=& - \textstyle \sum_{k=1}^N \big[ y_k \log P(\tau_{i_k} \succ \tau_{j_k}) + \nonumber \\
            && (1 - y_k) \log P(\tau_{j_k} \succ \tau_{i_k}) \big]
        \end{eqnarray}
        where $y_k$ is the query label from the human preference, with $y_k = 1$ if $\tau_{i_k}$ is preferred over $\tau_{j_k}$, and $y_k = 0$ otherwise.
        
        Using the estimated reward function $r(\tau)$, the agent learns a policy $\pi(a|s)$ within the reinforcement learning framework.
        The agent continuously receives human feedback as it takes new actions, updating the reward function $r(\tau)$ iteratively. This iterative process involves re-estimating the reward function based on new feedback data and refining the agent's policy.

\section{Proposed Framework}

    We propose Discriminability-Aware Policy-to-Policy Preference-Based Efficient Reinforcement Learning (DAPPER), a framework for query-efficient policy learning through easy-to-label queries.
    An overview is shown in \figref{fig:proposed_method}. Queries compare trajectories from different policies. Each query includes a preference label and a discriminability label assessing whether superiority can be determined.
    To enhance policy distinction, a discriminator is trained with the preference reward model to guide query sampling and promote diverse policies.
    
    The following sections describe the proposed framework.
    \chapref{chap:framework_components} introduces its main components.
    Using the reward functions described in \chapref{chap:reward_formulation}, DAPPER learns policies that reflect human preferences through the iterative execution of three steps: (1) Policy Learning, (2) Query Collection, and (3) Reward Update.
    Starting from \chapref{chap:policy_learning}, the subsequent sections provide details of each learning step. The overall procedure is summarized in the pseudo-code shown in \algref{algorithm}.

\begin{algorithm}[t]
    \small
    \SetKwData{Left}{left}\SetKwData{This}{this}\SetKwData{Up}{up}
    \SetKwFunction{Union}{Union}\SetKwFunction{FindCompress}{FindCompress}
    \SetKwInOut{Input}{input}\SetKwInOut{Output}{output}
    \caption{DAPPER}
    \label{algorithm}
    Set parameters described in \tabref{table:parameter_setting} \\
    Set policy dataset $\Pi$ and query dataset $\mathcal{D}$ \\
    \While{$ i = 1, 2, \ ..., \text{until converge}$}
    {
        Set current policy $\pi_{i}$ and rollout buffer $\mathcal{B}_i$ \\
        \For{$ j = 1, 2, \ ..., J$}
        {
            \For{$ e = 1, 2, \ ..., E$}
            {
                \For{$ t = 1, 2, \ ..., T$}
                {
                    Get observation $s_{t}$ from environment \\
                    Take action $a_t$ by $\pi_{i}(s_{t})$ \\
                    Calculate reward $r_t$ by \equref{eq:reward_concat_ours} \\
                    Push $(s_t, a_t, r_t)$ to $\mathcal{B}_i$ \\
                }
            }
            Update $\pi_i$ by RL update scheme (e.g. \cite{ConstraintPPO_1}) using $\mathcal{B}_i$ \\
        }
        Store learned policy to policy dataset $\pi_i \rightarrow \Pi$ \\
        \For{$ n = 1, 2, \ ..., N$}
        {
            Choose policy $\pi_{x}$ by \equref{query_policy_sampling} \\
            Generate trajecoties $\tau_{x}$, $\tau_{i}$ and conduct queries \\
            Store label $y$ to query dataset $(\tau_x, \tau_i, y) \rightarrow \mathcal{D}$
        }
        Update preference reward model $R^H$ by \equref{proposed_preference_loss} \\
        Update discriminator $D$ by \equref{proposed_disc_loss} \\
    }    
\end{algorithm}

\begin{table}[t]
    \small
    \vspace{1mm}
    \caption{
        Learning Parameters in Experiments
    }
    \label{table:parameter_setting}
    \vspace{-2mm}
    \begin{center}
        \begin{tabular}{@{}lp{5.5cm}ll@{}}
            \toprule
            \textbf{Para.} & \textbf{Meaning} & \textbf{Value} \\ 
            \midrule
            $\alpha$ & Coefficient of Query Sampling & $10^{-3}$ \\
            $\beta$ & Coefficient of Policy Exploration & $0.6$ \\
            $\gamma$ & Discount factor of RL & $0.99$ \\
            $N$ & Query number per iteration & $10$ \\
            $T$ & Number of steps per episode & $128$ \\
            $E$ & Number of episodes per one policy update cycle (Parallel environment number) & $128$ \\
            $J$ & Number of policy learning iterations to train $\pi_{i}$ & $300$ \\
            \bottomrule
        \end{tabular}
    \end{center}
\end{table}

    \subsection{Framework Components}
    \label{chap:framework_components}
        DAPPER consists of the following components.
        
        \textbf{Policy $\pi_i(s;\theta_i)$:}
            A policy used for learning human preferences is parameterized by $\theta_i$. Multiple policies with distinct parameters $\theta_i$ are trained for use in query comparisons, and the resulting policies are collected in the policy dataset $\Pi \ni (\pi_1, \pi_2, \dots, \pi_i, \dots)$.

        \textbf{Trajectory Generator $\mathcal{G}(\pi)$:}  
            A function that samples a trajectory $\tau$ from a policy $\pi$.  
            This function is used to make policy-based queries and also serves as input to the discriminator for comparing policies.  
            We define the generator as: $\mathcal{G}(\pi) = \tau \sim \pi,\quad \text{where } \tau = (s_0, a_0, s_1, \dots, a_{T-1}, s_T)$.
            Here, $T$ denotes the episode length.

        \textbf{Preference Reward Model $R^H(s,a;\phi)$:}  
            A reward function model for learning human preferences parameterized by $\phi$, defined identically to PbRL as described in \chapref{chap:pbrl}. Based on this reward function, human preferences are estimated as shown in \equref{eq:previous_preference}.

        \textbf{Discriminator $D(\mathcal{G}(\pi_{i}),\mathcal{G}(\pi_{j});\psi)$:}  
            A model that estimates the probability of whether preferences of policy pairs are discriminable by humans parameterized by $\psi$. By learning from previously trained policies, it functions as a reward estimation, ensuring that new policy remain distinguishable. Additionally, it is used to prioritize policy sampling when selecting policies for queries.

    \subsection{Reward Formulation}
    \label{chap:reward_formulation}
        Our framework proposes policy-based queries, in contrast to previous works that employ trajectory-based queries from a single policy as shown in \equref{eq:previous_preference}.
        Accordingly, we define human preference as follows:
        \begin{eqnarray}
            P(\pi_i \succ \pi_j) = \frac{\exp[R^H(\mathcal{G}(\pi_i))]}{\exp[R^H(\mathcal{G}(\pi_i))] + \exp[R^H(\mathcal{G}(\pi_j))]}.
        \end{eqnarray}
        
        In DAPPER, in addition to aligning with human preferences, the framework aims to learn policies whose preferences can be discriminated in queries. To this end, the following reward function $r$ is defined from preference reward model $R^H$ and discriminability reward $R^D$, as shown below:  
        \begin{eqnarray}
            \label{eq:reward_concat_ours}
            r =  (1-\beta) R^H + \beta R^D.
        \end{eqnarray}
        The hyperparameter $\beta$ trades off alignment with preference reward learned from human queries and policy diversity.

        Let $(s_t, a_t)$ be a one sample set of current step $t$ obtained from the interaction between the current policy $\pi_i$ and the environment, the discriminability reward $r^D$ is calculated from the discriminator $D$ as follows:  
        \begin{eqnarray}
            R^D(s_t, a_t) = \frac{1}{N'} \sum\nolimits_{\pi_x \in \Pi} D((\Bar{\tau_i}, s_t, a_t), \mathcal{G}(\pi_x)),
        \end{eqnarray}
        where $\Bar{\tau_i} = (s_0, a_0, s_1, \dots, a_{t-1}, s_{t-1})$ denotes the partial trajectory of policy $\pi_i$ up to step $t$, and $N'$ is the number of previously learned policies $(\pi_1, \pi_2, \dots, \pi_{i-1}) \in \Pi$ used for comparison.

    \subsection{Policy Learning}
        \label{chap:policy_learning}
        In this process, the policy $\pi_i$ for the current iteration $i$ is initialized and trained until the reward $r$ defined in \equref{eq:reward_concat_ours} converges.  
        Once training converges, the learned policy is added to the policy dataset $\Pi$ and used for query generation.

    \subsection{Query Collection}
        In this process, queries are generated from the learned policies, and labels are collected through the queries.  
        For query generation, DAPPER utilizes the discriminator $D$ to perform prioritized sampling of policies used in queries, aiming to enhance query discriminability. Specifically, one of the policy pairs is selected from the current policy $\pi_i$, which has high discriminability against past policies learned through $R^D$. The other policy $\pi_x$ is selected from the policy dataset $\Pi \ni (\pi_1, \pi_2 , \cdots, \pi_{i-1})$.  
        At this stage, the discriminability probability $D(\mathcal{G}(\pi_x),\mathcal{G}(\pi_i))$ between the current policy $\pi_i$ and a past policy $\pi_x \in \Pi$ is estimated. Based on this probability, policies for queries are sampled according to the sampling probability $p_x$ defined as follows:  
        \begin{eqnarray}
            \label{query_policy_sampling}
            p_x = \frac{\exp\big[\alpha D(\mathcal{G}(\pi_x),\mathcal{G}(\pi_i))\big]}{\sum_{j=1}^{N} \exp\big[\alpha D(\mathcal{G}(\pi_x),\mathcal{G}(\pi_j))\big]}
        \end{eqnarray}
        where $\alpha$ is a temperature parameter that controls sampling sharpness: higher values of $\alpha$ favor more discriminable policies, while lower values result in more uniform sampling. Sampling is performed without duplication of any policy pairs already present in the query dataset $\mathcal{D}$.

        Then, using the sampled policies, a state-action trajectory $\tau$ is generated from the environment, and the label $y = P(\pi_x \succ \pi_i)$ is collected.  
        In this process, the label $y$ takes values from $\{0, 0.5, 1\}$, where $y = 1$ indicates that $\pi_x$ is preferred over $\pi_i$, $y = 0$ indicates that $\pi_i$ is preferred over $\pi_x$, and $y = 0.5$ signifies that both policies are considered equally preferable or that ranking them is difficult.  
        The obtained labeled data $(\pi_x, \pi_i, y)$ is added to the query dataset $\mathcal{D}$.

    \subsection{Reward Update}
    
        \subsubsection{Execution Flow}
            Using the query dataset $\mathcal{D}$, the preference reward model $R^H$ for calculating $r^H$ and the discriminator $D$ for calculating $r^D$ are updated.  
            After the models $R^H$ and $D$ have converged, they are utilized in training the next iteration's policy $\pi_{i+1}$. This process enables learning policies that reflect human preferences from queries while also promoting the acquisition of policies that remain distinguishable from past policies $\Pi$.

        \subsubsection{Preference Reward}
            Using the separable label dataset $y \in \{0,1\}$ from the query dataset $\mathcal{D}$, the preference reward model $R^H$ is updated.
            The loss function is defined as follows:  
            \begin{eqnarray}
                \label{proposed_preference_loss}
                \mathcal{L}_{\phi} &=& -\mathbb{E}_{(\pi_i,\pi_x, y) \sim \mathcal{D}} \Big[ y \log P(\pi_{x} \succ \pi_{i}) \nonumber \\
                && + (1 - y) \log P(\pi_{i} \succ \pi_{x}) \Big].
            \end{eqnarray}

        \subsubsection{Discriminability Reward}
            The discriminator $D$ is trained on the query dataset $\mathcal{D}$, where the label $y$ can take inseparable values $y \in \{0.5, \text{other}\}$. 
            The loss function is defined as follows:  
            \begin{eqnarray}
                \label{proposed_disc_loss}
                \mathcal{L}_{\psi} &=& -\mathbb{E}_{(\pi_i,\pi_x, y) \sim \mathcal{D}} \Big[ y \log D(\mathcal{G}(\pi_{x}), \mathcal{G}(\pi_{i})) \nonumber \\
                && \quad + \; (1-y) \log \big(1 - D(\mathcal{G}(\pi_{x}), \mathcal{G}(\pi_{i}))\big) \Big],
            \end{eqnarray}
            where $D$ is defined as a symmetric function, i.e., $D(\pi_x,\pi_i) = D(\pi_i,\pi_x)$.  
            This formulation is equivalent to the standard binary cross-entropy loss, encouraging the discriminator $D$ to assign high probability when two policy embeddings are distinguishable, and low probability otherwise.

\begin{table}[t]
    \small
    \vspace{1mm}
    \caption{
        Constraint parameters: 
        Following the definitions in \chapref{sec:policy_constraint}, the constraint target parameters for each task are defined. 
        These parameters are estimated by feature extractor $f$ from state-action trajectories $\tau$. The threshold is set to a common value of $d = 10$ for all parameters.
    }
    \label{table:policy_constraint}
    \vspace{-2mm}
    \begin{center}
        \begin{tabular}{lcc}
            \toprule
            \textbf{Parameter} & \textbf{Target: $x^*$} & \textbf{Marge: $g$} \\ 
            \midrule
            Body Velocity of X Direction [m/s] & 1 & 0.2 \\
            Body Velocity of Y Direction [m/s] & 0 & 0.2 \\
            Body Rotation of Z Axis [degree] & 0 & 10 \\
            Collisions on Bodies and Legs & 0 & 0 \\
            \bottomrule
        \end{tabular}
    \end{center}
\end{table}

\begin{table*}[t]
    \small
    \vspace{1mm}
    \caption{
        Terms of feature extraction:
        The second and third columns show the parameter ranges, while the fourth column presents the target feature values $f^*$. The remaining columns indicate the walking patterns and the features used in each pattern, where a checkmark means the feature follows $f^*$, and a numerical value means it follows the specified value.
        ``Gait Cycle'' and ``Phase Difference'' are calculated from the timing of foot-ground contact.
        ``FL'' (Front Left), ``FR'' (Front Right), ``RL'' (Rear Left), and ``RR'' (Rear Right) refer to the respective legs of a quadrupedal robot.
        ``Gait Cycle'' is the duration of a foot's contact cycle, while ``Phase Difference'' denotes the delay between two legs' contact timings.
        Both are updated at each foot contact, with the most recent value used at step $t$.
        ``Phase Difference'' is given as the delay ratio to the current gait cycle. ``Body Height'' and ``Body Incline Angle'' are computed at each step $t$.
    }
    \label{table:feature_extractor}
    \vspace{-2mm}
    \begin{center}
        \begin{tabular}{lcccccccc}
            \toprule
            \textbf{Parameter} & \textbf{Min} & \textbf{Max} & \textbf{Target $f^*$} & \textbf{Posture} & \textbf{Trot Only} & \textbf{Normal} & \textbf{Crawl} \\ 
            \midrule
            Body Height [cm] & 25 & 65 & 45 & 45 & - & 55 & 30 \\
            Body Incline Angle [degree] & -10 & 10 & 0 & \checkmark & - & \checkmark & \checkmark \\
            Gait Cycle (FL) [s] & 0.2 & 0.4 & 0.5 & - & \checkmark & \checkmark & \checkmark \\
            Phase Difference (FL-FR) [\%] & 0 & 100 & 50 & - & \checkmark & \checkmark & \checkmark \\
            Phase Difference (FL-RL) [\%] & 0 & 100 & 50& - & \checkmark & \checkmark & \checkmark \\
            Phase Difference (FL-RR) [\%] & 0 & 100 & 0 & - & \checkmark & \checkmark & \checkmark \\
            \bottomrule
        \end{tabular}
    \end{center}
\end{table*}

\section{Experiments}
    The following experiments investigate whether the proposed framework can achieve higher query efficiency than existing methods by considering preference discriminability.
    In comprehensive comparative experiments, it is impractical for an actual human to manually label all queries due to the enormous time and human cost.
    To address this, we define a simulated annotator model and perform automatic labeling. We first formulate the simulated annotator's preference model and preference discriminability as mathematical expressions.
    Subsequently, we conduct a human subject experiment to validate the simulated annotator model by examining whether actual human responses to trajectory comparison queries can be explained by the preference discriminability defined in the model (at \chapref{ex_humanQuery}).
    From the experimental results, we quantitatively determine an approximate threshold for preference discriminability and define multiple levels around it.
    These thresholds are then used in the subsequent PbRL experiments (from \chapref{ex_queryNum}) to automate query labeling with the simulated annotator model, enabling efficient evaluation of the proposed framework.

    In our experiments, we aim to address the following research questions:
    \begin{itemize}
        \item To what extent does the preference discriminability defined in the simulated annotator model explain actual human responses to trajectory comparison queries? (\chapref{ex_humanQuery})
        \item Does our framework improve query efficiency compared to previous methods? (\chapref{ex_queryNum})
        \item How does our framework perform when the threshold of preference discriminability is set substantially above or below the range observed in actual human responses? (\chapref{ex_threshold})
        \item How does our framework affect computational time efficiency? (\chapref{ex_calculationTime})
        \item How does hyperparameter selection affect query efficiency and preference discriminability in our framework? (\chapref{ex_methodParameter})
        \item Can our framework effectively learn from VLM-based preference labels, which rely on world knowledge aggregated from large-scale data, when used as substitutes for human queries that may include labeling errors or indistinguishable cases? (\chapref{ex_VLM})
        \item Can the proposed method maintain its effectiveness as the number of feature dimensions that parametrize the target robot behaviors increases? (\chapref{ex_task_scalability})
        \item Is our framework applicable to sim-to-real policy transfer? (\chapref{ex_sim2real})
    \end{itemize}

\begin{figure*}[t]
    \hspace{-5mm}
    \centering
    \includegraphics[width=1.85\columnwidth]{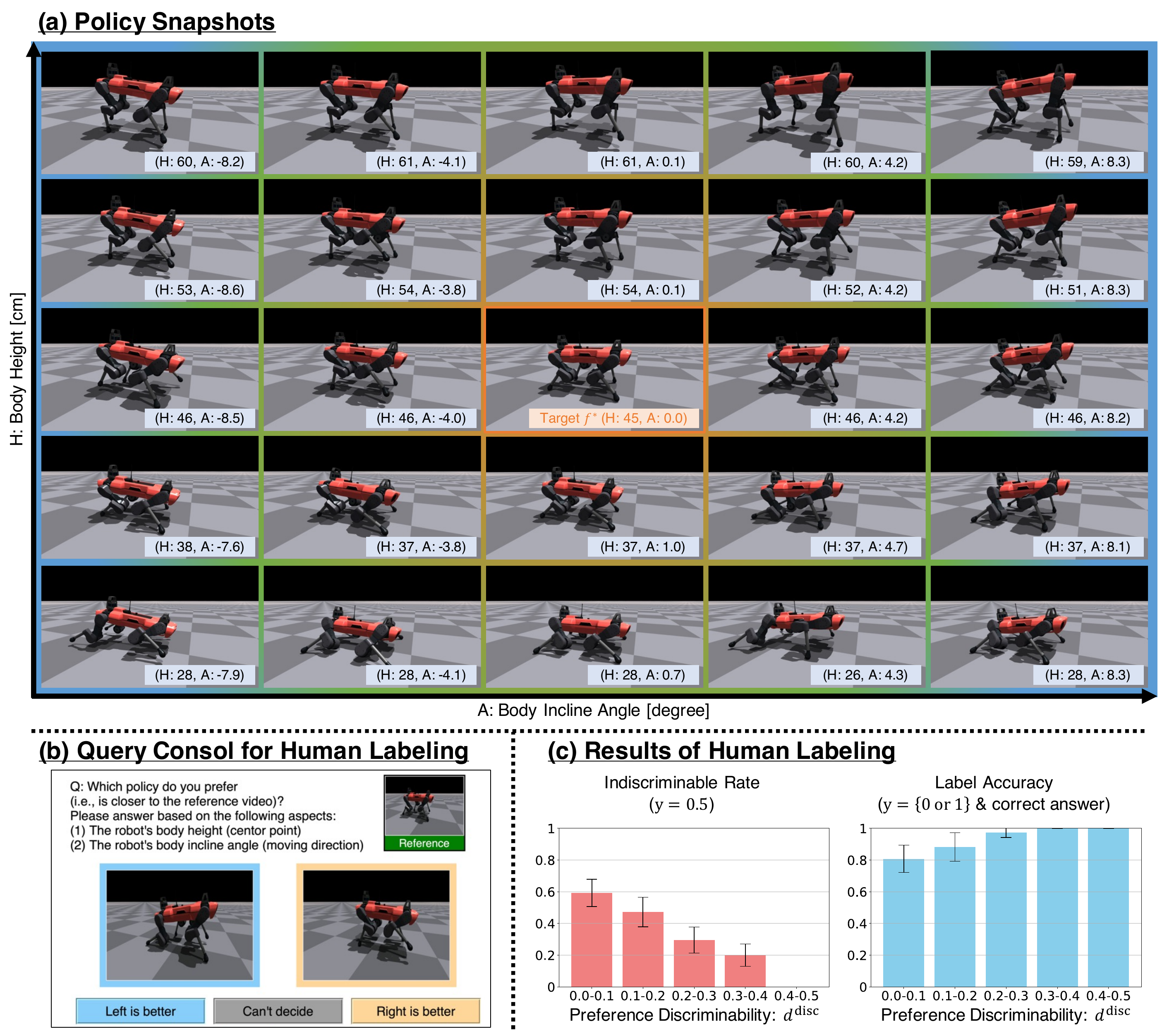}
    \vspace{-1mm}
    \caption{
        Analysis of actual human query responses for determining the threshold of preference discriminability $d^{\text{disc}}$ in the simulated annotator model:
        A questionnaire was conducted using a console system, as shown in (b).
        At the top of the screen, a reference video was displayed, showing a policy that matched a predefined target behavior.
        Below the reference video, two candidate policy videos were shown for comparison.
        Participants were asked to choose the trajectory closer to the reference by selecting one of three options: ``Left is better'', ``Right is better'', or ``Can't decide''.
        The candidate policies were prepared by exhaustively combining the listed values of $\text{Body Height [cm]} \in \{29, 37, 45, 53, 61\}$ and $\text{Body Incline Angle [degree]} \in \{-8, -4, 0, 4, 8\}$.
        For each combination of these parameters, a policy was trained using DAPPER under an idealized setting where $d^{\text{disc}}=0$, by optimizing the behavior to match the specified Body Height and Body Incline Angle values until convergence.
        Participants were instructed to compare only two features defined by $f^*$: Body Height and Body Incline Angle. Gait patterns were excluded. They could prioritize the features freely, but were told to consider both.
        Policy pairs were sampled from (a), and 105 predefined queries were presented sequentially.
        The queries were selected to ensure diversity in preference discriminability $d^{\text{disc}}$.
        All participants evaluated the same set of queries, and the order of the queries was randomized for each participant.
        The results shown in (c) represent the mean and standard deviation across five participants.
    }
    \label{fig:ex4}
    \vspace{-3mm}
\end{figure*}

\subsection{Common Settings}
    All experiments are conducted using the ANYmal quadrupedal robot and the \textit{legged\_gym} RL-framework based on the IsaacGym physics simulator \cite{legged_gym}. 
    The trained policy is evaluated at \SI{50}{Hz} predicting target joint positions, which are tracked by per joint PID controller running at \SI{400}{Hz}. 
    In all experiments, the learning parameter settings follow those of previous works, including simulation parameters, specifically state-action definition and network architectures \cite{legged_gym}.
    Policy learning utilized Lagrangian Proximal Policy Optimization (LPPO) \cite{ConstraintPPO_1,ConstraintPPO_2} with parallel environments, and all experiments are conducted using a Nvidia GeForce RTX 4090 GPU.
    To stabilize the $R^D$ signal particularly during the early training phase, we introduced Monte Carlo dropout \cite{mc_dropout} into the discriminator $D$.
    Furthermore, to prevent the discriminator $D$ from overfitting to earlier policies when continuously updated across iterations, their parameters were reset and retrained from scratch at each iteration.

    \subsubsection{Feature Extraction}
        To quantify the degree of alignment, we utilize the policy feature representation $f(s,a)$ extracted by a feature extractor introduced in previous PbRL works \cite{reward_encoding_origin, feature_extraction_pbrl}. In the feature space defined by $f$, we specify a target feature $f^*$ that represents the desired behavior in accordance with human preferences. The target features are listed in \tabref{table:feature_extractor}.
        In this paper, feature extraction is also performed when comparing different policies. In such cases, trajectory $\tau$ is generated by a policy $\pi$, then the average feature representation is calculated as $f(\pi) = \frac{1}{T} \textstyle \sum_{(s, a) \in \tau} f(s, a)$, where $T$ denotes the episode length.

    \subsubsection{Constraint Terms of LPPO}
        \label{sec:policy_constraint}
        This paper adopts LPPO to learn a policy that satisfies constraints while suppressing unrealistic behaviors that may hinder accurate query labeling. Constraint term is defined by indicator function $\mathbb{I}$, allowing the cost function $c$ to be expressed as follows: $c(s_t, a_t) = \mathbb{I}\left[|x_t - x^*| > g\right]$, where $x_t$ denotes a current value of constraint target parameter, $x^*$ is the target value, and $g$ is the permissible margin of error. This formulation penalizes each timestep in which the deviation from the target exceeds the margin $g$.
        To enforce the constraint over an episode of length $T$, LPPO defines an threshold $d$ on the total number of penalty-inducing steps: $\textstyle \sum_{t=1}^{T} c(s_t, a_t) \leq d$. This constraint ensures that the number of steps in which the system deviates significantly from the desired behavior remains within an acceptable bound.
        The constraint terms are listed in \tabref{table:policy_constraint}.

    \subsubsection{Labeling Queries from Simulated Annotator Model}
        To reduce the human cost associated with large-scale evaluation, we employ a simulated annotator model.
        Specifically, we define a simulated annotator's preference model and preference discriminability as mathematical expressions, enabling automatic labeling.
        The simulated annotator assigns labels $y \in {0, 0.5, 1}$ based on the extracted policy features $f(\pi)$ and a predefined target feature $f^*$.
        All feature values are normalized to the range $[0,1]$ according to the parameter specifications listed in \tabref{table:feature_extractor}.

        To estimate preference and discriminability labels, we define a unified discriminability metric based on the difference in feature errors between two trajectories $\tau_i$ and $\tau_j$ with respect to a target feature $f^*$. Specifically, we use the following scale-normalized L1-based distance:  
        \begin{subequations} \label{eq:d}
            \begin{flalign}
                & d^{\text{pref}}(\tau, f^*) = \frac{1}{\sqrt{|f|}} \textstyle \sum_{k=1}^{|f|} |f_k(\tau) - f^*_k|, \label{eq:d_pref} \\
                & d^{\text{disc}}(\tau_i, \tau_j, f^*)= |d^{\text{pref}}(f(\tau_i), f^*) - d^{\text{pref}}(f(\tau_j), f^*)| , \label{eq:d_disc}
            \end{flalign}
        \end{subequations}
        where $|f|$ denotes the number of feature dimensions.
        Here, the trajectory with the smaller feature error $d^{\text{pref}}$ is labeled as more preferred ($y \in \{0,1\}$), while the discriminability label ($y \in \{0.5, \text{other}\}$) is determined by whether preference discriminability $d^{\text{disc}}$ exceeds a predefined threshold. If the value is above the threshold, the query is considered distinguishable; otherwise, it is treated as indistinguishable.

\begin{figure*}[t]
    \centering
    \includegraphics[width=1.55\columnwidth]{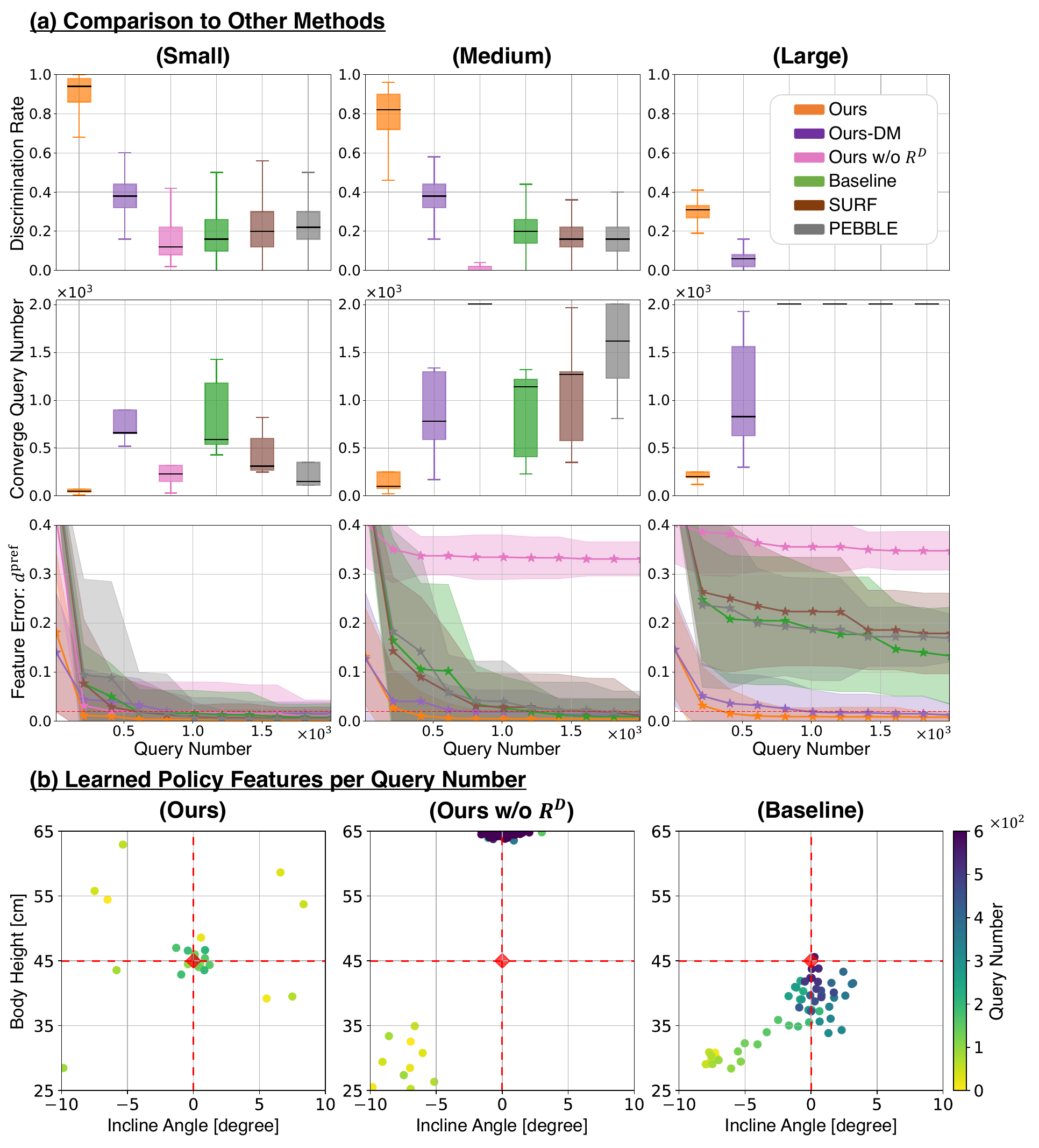}
    \vspace{-1mm}
    \caption{
        Evaluation of query efficiency:
        \textbf{(a)} Comparison to previous works.
        \textbf{Top:} Discrimination rate of queries.
        \textbf{Middle:} Converge query number for reaching the feature error threshold ($d^{\text{pref}} < 0.02$), empirically defined by the convergence values of each method. If not converged within $2 \times 10^3$ queries, the value is clipped at $2 \times 10^3$. The threshold is shown as a red dashed line in the bottom plot.
        \textbf{Bottom:} Minimum values of feature error $d^{\text{pref}}$ achieved by the learned policies. ``Small'', ``Medium'', and ``Large'' denote levels of preference discriminability $d^{\text{disc}}$. Each curve shows the mean and standard deviation over five experiments.
        \textbf{(b)} Learned policy features per query number.
        This experiments use $d^{\text{disc}}$ = ``Medium''.
        The target feature $f^*$ is located at the intersection of the red dashed lines. The number of plotted points corresponds to the queries collected until the feature error falls below the threshold ($d^{\text{pref}} < 0.02$), indicated by the red frame.
    }
    \label{fig:ex1}
    \vspace{-3mm}
\end{figure*}

\subsection{Comparing Methods}
\label{comparing_methods}

    We compared the following methods to verify the effectiveness of the proposed method:
    \begin{itemize}
        \item Baseline \cite{pbrl}: A method that generates queries from pairs of state-action trajectories obtained from different episodes using a single policy.  
        \item SURF \cite{PbRL_auto_label}: Same as the Baseline, queries are generated from a single policy. However, this method automatically labels queries when the preference values calculated from the learned reward function are sufficiently high, thereby increasing the number of training samples.
        \item PEBBLE \cite{pebble}: Extends the Baseline with unsupervised pre-training and off-policy learning. The agent explores with intrinsic rewards maximizing state entropy to generate diverse behaviors before learing from queries. Past reward values are re-calculated and utilized after each reward model update, enabling sample-efficient policy learning.
    \end{itemize}

    We also verified the following two ablation methods of our framework:  
    \begin{itemize}
        \item Ours w/o $R^D$: Same as the proposed framework, queries are generated from pairs of state-action trajectories obtained from different policies. However, the reward function $R^D$, which encourages acquiring policies with distinct features, is not used.  
        \item Ours with Distance Maximization (Ours-DM): This is the proposed framework with different $R^D$ which is calculated based on the feature distance from previously learned policies using the following equation: ${R^{D}}' = \frac{-1}{N'} \sum_{\pi_x \sim \Pi} |f(\pi_i) - f(\pi_x)|$. This setting is included to evaluate whether simply maximizing the distance between policies is more effective than Ours.
    \end{itemize}

\subsection{Analysis of Actual-Human Preference Discriminability}
\label{ex_humanQuery}
    \subsubsection{Settings} 
        This paper proposes a framework for generating queries that are easier for humans to distinguish, based on preference discriminability.
        This section examines whether actual humans can distinguish similar queries and how their ability declines with increasing similarity.
        To control for individual preference variability, we define a target feature $f^*$ and treat the policy closer to $f^*$ as the correct answer.
        We then analyze the proportion of indistinguishable responses ($y = 0.5$) as a function of $d^{\text{disc}}$ to quantify actual-human preference discriminability.
        Additionally, we evaluate the accuracy of query labels when participants express a clear preference ($y \neq 0.5$).
        This experiment uses the ``Posture'' feature values defined in \tabref{table:feature_extractor}.

    \subsubsection{Results}
        The experimental results are shown in \figref{fig:ex4}.  
        As seen in \figref{fig:ex4}(c), when $d^{\text{disc}}$ is low, around \SI{60}{\%} of queries were judged as indistinguishable. As $d^{\text{disc}}$ increases, distinguishability improves, reaching \SI{100}{\%} for all participants when $d^{\text{disc}} > 0.4$.
        Label accuracy for distinguishable queries exceeded \SI{80}{\%} at all $d^{\text{disc}}$ levels, and reached \SI{100}{\%} when $d^{\text{disc}} > 0.3$.
        These results suggest that $d^{\text{disc}} > 0.4$ ensures consistent discrimination, while lower values depend on individual ability. Based on this, we define three threshold levels as criteria for automatic labeling in the subsequent sections: ``Small'' ($0.2$), ``Medium'' ($0.3$), and ``Large'' ($0.4$).

\begin{figure}[t]
    \hspace{-4mm}
    \centering
    \includegraphics[width=0.99\columnwidth]{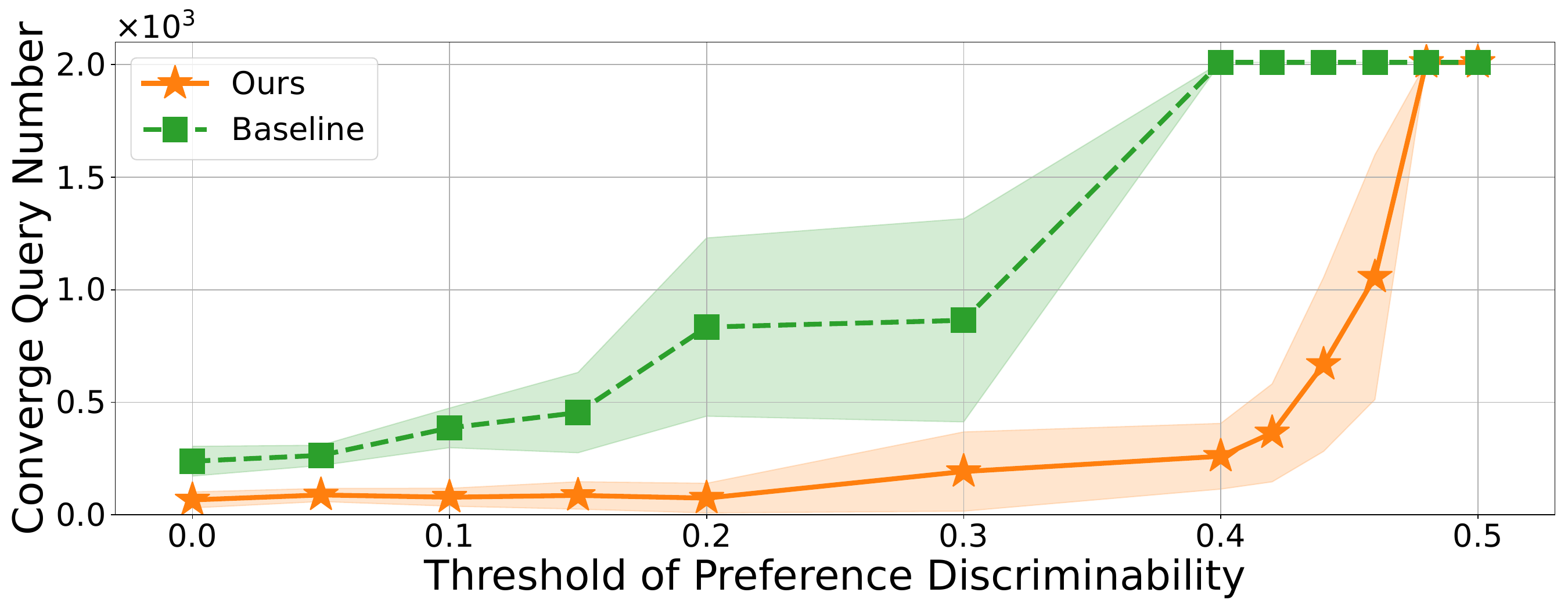}
    \caption{
        Performance comparison across different preference discriminability thresholds:
        The convergence query number denotes the point at which the learned policy reaches the feature error threshold ($d^{\text{pref}} < 0.02$).
        If convergence is not achieved within $2 \times 10^3$ queries, the value is clipped at $2 \times 10^3$.
        Each curve shows the mean and standard deviation over five experiments.
    }
    \label{fig:threshold}
\end{figure}

\subsection{Comparison of Query Efficiency with Other Methods}
\label{ex_queryNum}
    \subsubsection{Settings} 

        The proposed framework improves query efficiency by utilizing queries from multiple policy trajectories instead of a single policy trajectory, thereby enhancing query discriminability to improve query efficiency.  
        To verify its effectiveness, we compare the query discrimination rate and the number of queries required to reach the predefined target feature $f^*$ (as shown in \tabref{table:feature_extractor}) against the other methods described in \chapref{comparing_methods}.
        We evaluate all methods across three levels of preference discriminability $d^{\text{disc}}$ to examine their robustness under increasingly difficult query discrimination conditions.
        This experiment uses target feature values of ``Posture'' as defined in \tabref{table:feature_extractor}.

    \subsubsection{Results}
        As shown in \figref{fig:ex1}(a), Ours (DAPPER) achieves the target features with the fewest number of queries.  
        When queries become harder to distinguish due to increased preference discriminability $d^{\text{disc}}$, the number of required queries increases. 
        In particular, when $d^{\text{disc}}$ is ``Large'', Ours requires the fewest queries; by contrast, the ablation variant requires over three times more queries, and the trajectory-based method fails to learn the target features within the given number of queries.  
        Ours consistently achieves the highest discrimination rate, and under the ``Large'' condition, only Ours and Ours-DM are able to generate a sufficient number of discriminable queries for learning a policy that closes the target features.

        As shown in \figref{fig:ex1}(b), the Baseline suffers from policy bias that constrains exploration to the features of previously learned policies and slows progress toward the target features $f^*$.
        In contrast, Ours w/o $R^D$ initially learns policies with similar feature representations because it lacks $R^D$, but retraining policies from scratch at each reward model update reduces the influence of policy bias and occasionally allows larger jumps to new feature regions.
        However, without the discriminability reward $R^D$ to actively encourage exploration, the policy tends to converge to local optima due to lack of discriminable queries.
        Meanwhile, Ours (i.e., with $R^D$) integrates training-from-scratch with the discriminability reward, consistently acquiring diverse and distinguishable policies that successfully converge toward the target feature value $f^*$.

\begin{figure}[t]
    \hspace{-4mm}
    \centering
    \includegraphics[width=0.99\columnwidth]{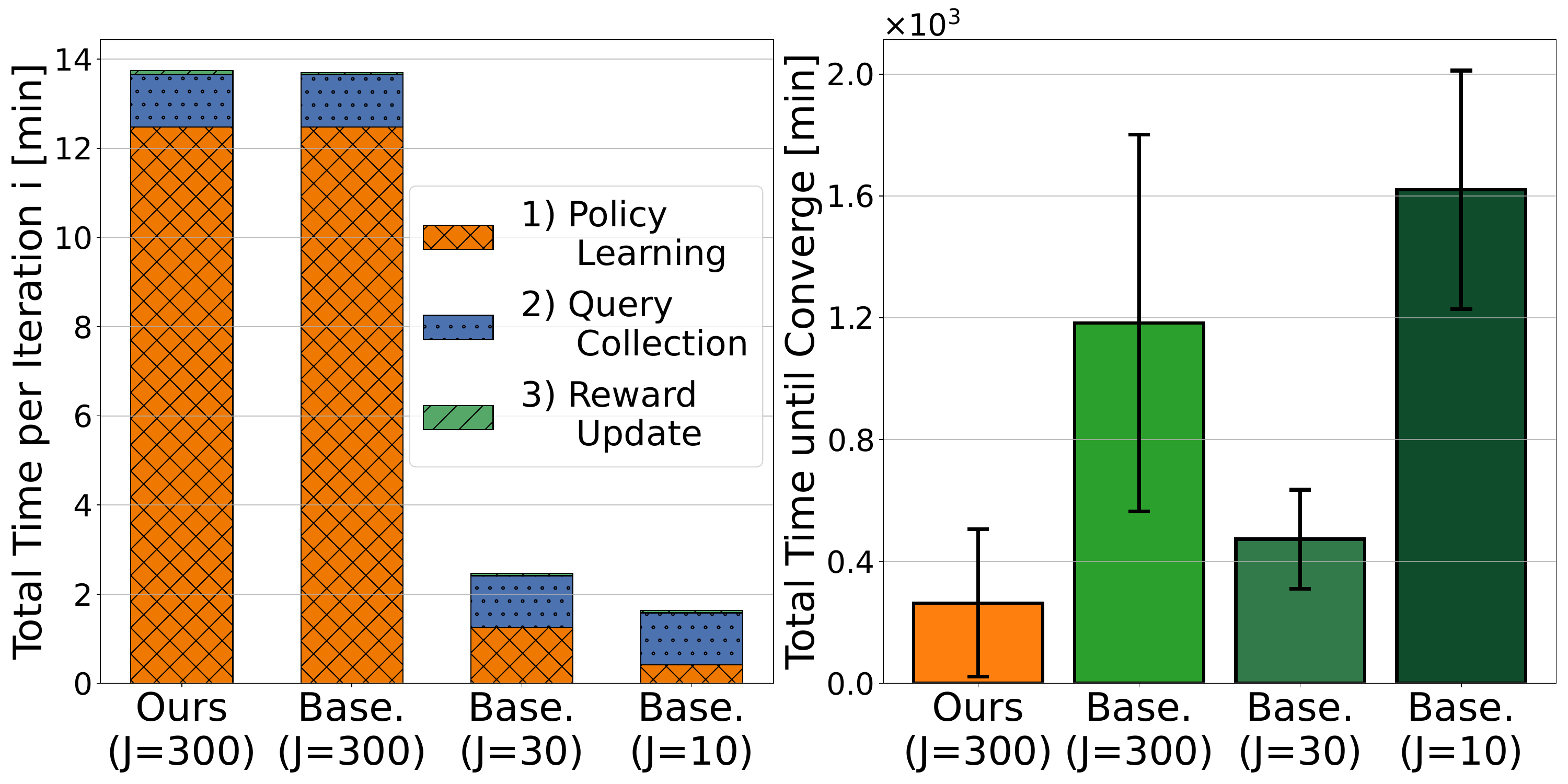}
    \caption{
        Analysis of computational time:
        \textbf{(Left)} Total computational time per iteration $i$. As shown in \figref{fig:proposed_method} and \algref{algorithm}, each iteration $i$ consists of three processes: ``1) Policy Learning'', ``2) Query Collection'', and ``3) Reward Update''.  
        In this plot, the stacked bar represents the total computational time, computed as the sum of the mean time spent on each process (averaged across the five trials). 
        Note that the ``3) Reward Update'' process takes only a few seconds and is barely visible in the plot.
        \textbf{(Right)} Total computational time until convergence, defined as the point where the learned policy's feature error satisfies the threshold ($d^{\text{pref}} < 0.02$).
        The bars show the mean time to converge, and the error bars denote the standard deviation across the five trials.
        As shown in \algref{algorithm}, the parameter $J$ represents the number of policy learning iterations used to train $\pi_i$ in ``1) Policy Learning'' within each cycle.
        The computation time for ``2) Query Collection'' is estimated at about 70 seconds based on \figref{fig:ex4}, where one human response took around 7 seconds on average and each iteration included $N = 10$ queries.
    }
    \label{fig:calculation_time}
\end{figure}

\subsection{Analysis of Sensitivity to Preference Discriminability Threshold}
\label{ex_threshold}
    \subsubsection{Settings} 
        As shown in \figref{fig:ex1}, our framework demonstrates improved query efficiency under preference discriminability thresholds corresponding to the range observed in actual human responses.
        In this section, we investigate how the framework behaves when the threshold is set beyond this human-observed range, either higher or lower, to observe its behavior under more challenging or relaxed discriminability conditions.
        For comparison, the Baseline is used to examine how our framework behaves differently under these varying threshold conditions.
        This experiment uses the target feature value of ``Posture''.

    \subsubsection{Results}
        As shown in \figref{fig:threshold}, Ours consistently outperforms the Baseline across a wide and fine-grained range of thresholds.
        The results also show that the performance gap narrows when the preference discriminability threshold is very low. However, even when the threshold is set to zero, meaning that all queries are distinguishable, the Baseline still performs worse than Ours because policy bias prevents it from learning substantially different behaviors within a small number of iterations $i$.
        Conversely, as the threshold increases and discriminable queries become scarce, policy learning becomes increasingly difficult, and Ours ultimately fails to learn policies.

\subsection{Analysis of Computational Time Efficiency}
\label{ex_calculationTime}
    \subsubsection{Settings} 
        As shown in \figref{fig:ex1}, this paper demonstrates improvements in preference discriminability and the resulting query efficiency. However, achieving these improvements introduces additional computational components. Specifically, a new policy is retrained from scratch at each iteration $i$ to avoid policy bias, and a discriminator $D$ is introduced as an additional component to estimate preference discriminability.
        In this experiment, we evaluate whether the gain in query efficiency achieved by our framework outweighs the additional computational cost in terms of total computation time until policy-learning converge, by examining whether it remains more time-efficient than the Baseline, which does not include these additional components.
        We evaluate the Baseline under different numbers of policy learning iterations ($J \in \{300, 30, 10\}$), since the Baseline does not initialize the policy at each iteration $i$ and can continue learning with fewer policy learning steps.
        This experiment uses target feature values of ``Posture'' under the ``Medium'' preference discriminability condition.

    \subsubsection{Results}
        The results are shown in \figref{fig:calculation_time}.
        Regarding total time per iteration $i$, the ``1) Policy Learning'' process takes about 12--13 minutes, accounting for most of the total 13--14 minutes.
        In contrast, the ``3) Reward Update'' requires only a few seconds, having a negligible effect on the total time.
        Reducing the number of policy learning iterations $J$ naturally decreases the overall computation time per iteration $i$.
        Regarding the total time until convergence, with $J = 30$, the Baseline reduces the overall time to roughly \SI{40}{\%} of that with $J = 300$, but it is still approximately twice as long as Ours.
        However, when $J$ is reduced to $10$, the time for ``1) Policy Learning'' becomes smaller, but ``2) Query Collection'' dominates, and more iterations are required for convergence, resulting in longer total training time.
        Overall, even when accounting for retraining from scratch and the additional discriminator $D$, Ours demonstrates superior time efficiency compared to the Baseline.

\subsection{Influence of Different Query Exploration/Sampling}
\label{ex_methodParameter}
    \subsubsection{Settings}
        As shown in \equref{eq:reward_concat_ours}, the reward in DAPPER is adjusted by setting the ratio $\beta$ between $r^D$ and $r^H$ to balance policy discriminability and convergence toward the target feature values.  
        Additionally, the policy sampling from the policy dataset $\Pi$ for query generation also affects discriminability performance (\equref{query_policy_sampling}).  
        Therefore, this section comprehensively compares these parameters and evaluates their influence on convergence to target features. 
        This experiment uses target feature values of ``Posture'' as defined in \tabref{table:feature_extractor}.

    \subsubsection{Results}
\begin{figure}[t]
    \hspace{-4mm}
    \centering
    \includegraphics[width=0.99\columnwidth]{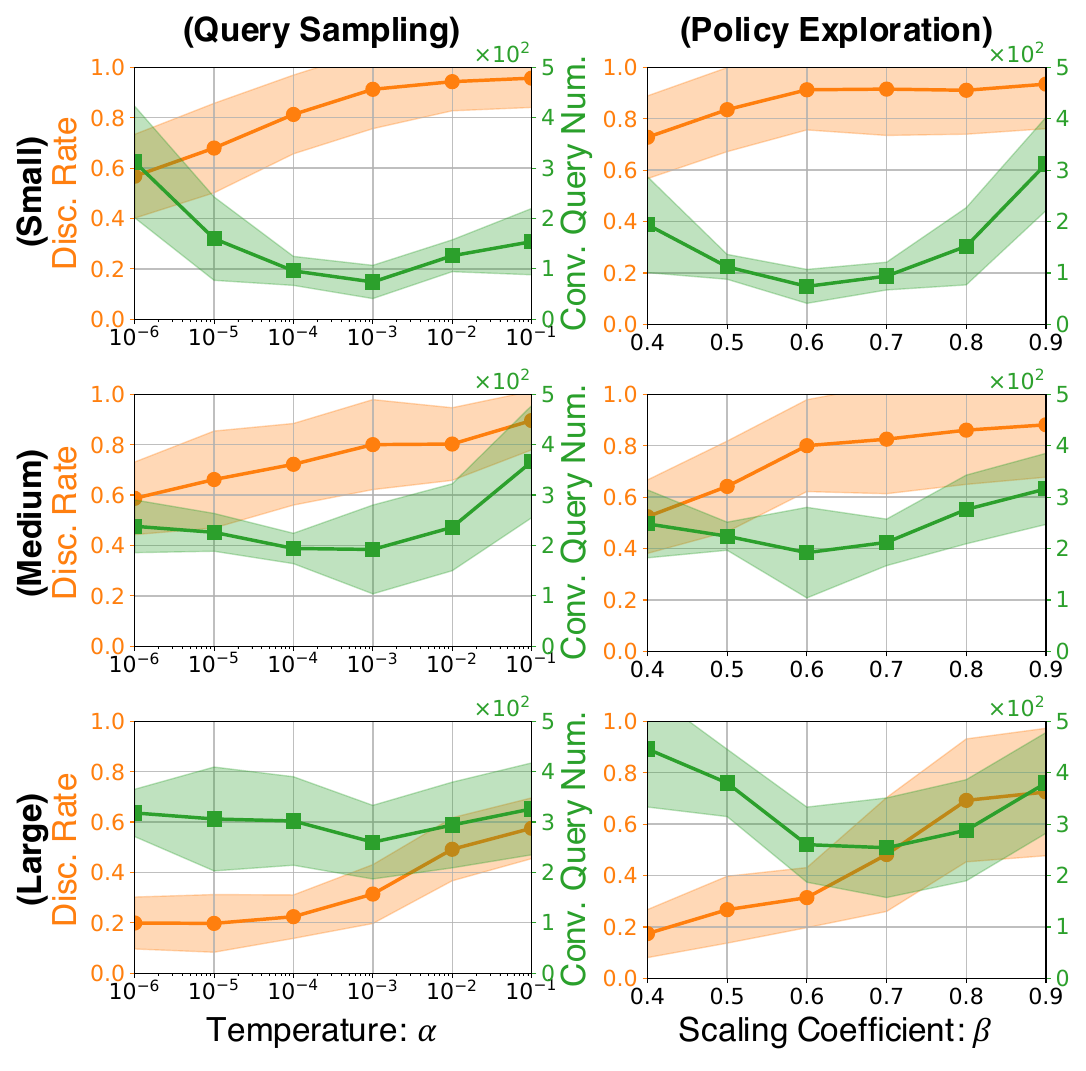}
    \vspace{-1mm}
    \caption{
        Performance comparison of DAPPER components:
        \textbf{(Left)} different temperature patterns of query sampling.
        \textbf{(Right)} different scaling coefficient patterns of policy exploration.
        ``Small'', ``Medium'', and ``Large'' denote levels of preference discriminability $d^{\text{disc}}$.
        Each curve plots the mean and standard deviation over five experiments. 
    }
    \label{fig:ex3}
\end{figure}

        The results are shown in \figref{fig:ex3}.
        Regarding query sampling, a higher temperature coefficient $\alpha$ increases the probability of selecting policies with high discriminability, thereby improving the overall discriminability rate.
        However, a large $\alpha$ results in biased sampling, as only easily distinguishable policies are selected by the discriminator $D$.
        Moderately reducing $\alpha$ improves query efficiency by encouraging diverse queries, which better support learning the preference reward $R^H$.
        Regarding policy exploration, a higher weight $\beta$ promotes broader exploration, which is expected to facilitate reaching the optimal policy.
        However, excessive exploration degrades convergence, as also observed in the results of Ours-DM in \chapref{ex_queryNum}.
        Since $\alpha = 10^{-3}$ and $\beta = 0.6$ consistently yield good performance across different $d^{\text{disc}}$ settings, we use them in all experiments.

\begin{figure}[t]
    \centering
    \includegraphics[width=0.95\columnwidth]{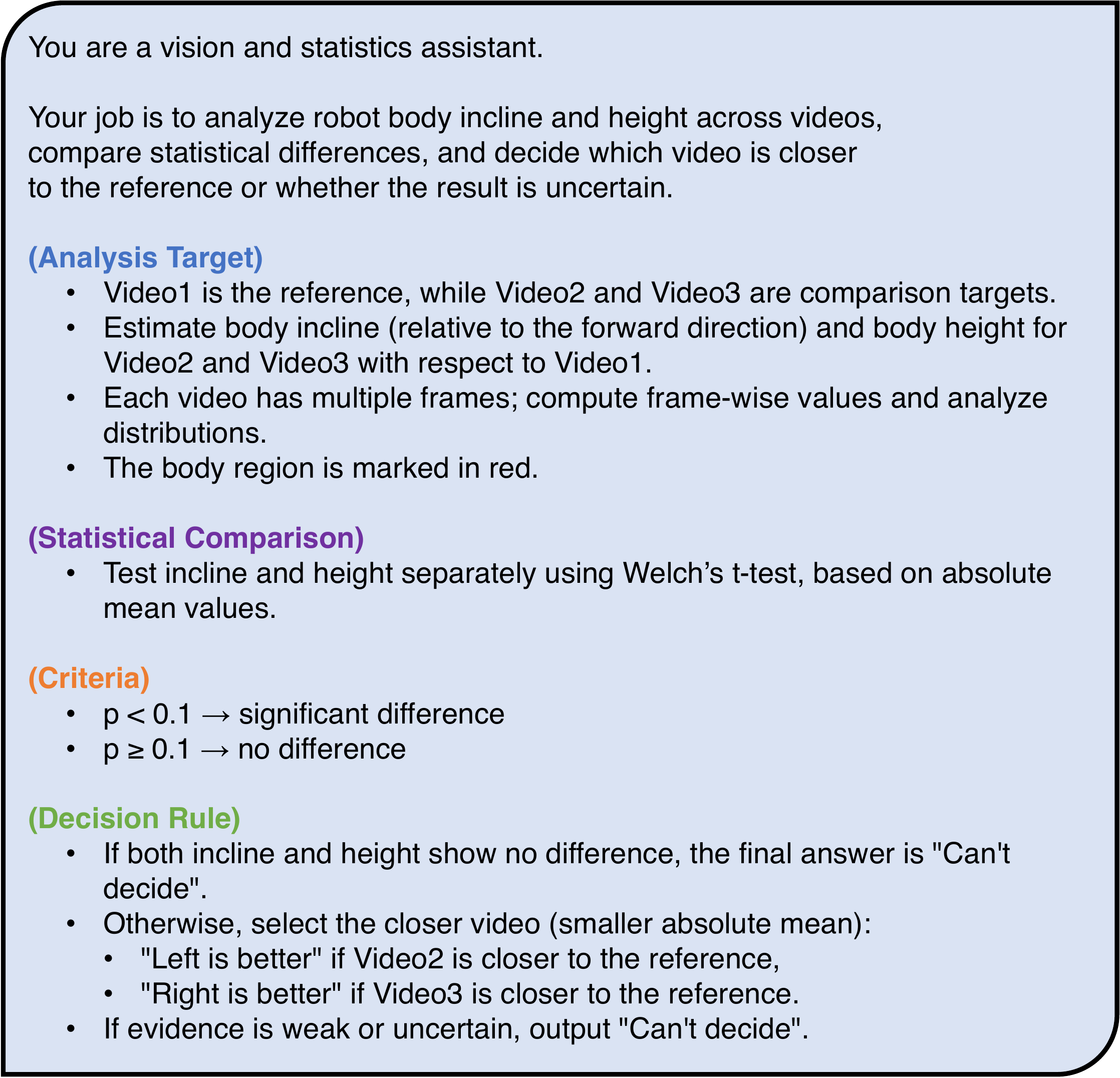}
    \caption{
        Prompt for VLM-based preference labeling:
        We used OpenAI GPT-5 model with the instruction text shown in the figure to generate preference labels. Three videos were used in each comparison: two candidate videos and one reference video. Each video is 3 seconds long at 30 Hz, from which three frames were sampled at 0.5, 1.5, and 2.5 seconds and downscaled from the original resolution of 1,600×900 pixels (the same resolution as in \figref{fig:ex4}) to 800×450 pixels to reduce computational and resource costs. These processed frames, together with the textual instructions, were used as inputs. The model was then asked to output one of three labels: ``Left is better'', ``Right is better'', or ``Can't decide''.
    }
    \label{fig:gpt_prompt}
\end{figure}

\subsection{Applicability of VLM-Based Preference Labels as Human Query Substitutes}
\label{ex_VLM}

    \subsubsection{Settings}  
        Vision-language models (VLMs) provide an attractive alternative to human annotators for generating preference labels \cite{vlm_asReward,vlm_spatial}. Using VLMs instead of humans could significantly reduce annotation cost, but it also introduces challenges such as noisy or overconfident predictions. Highlighting this alternative is particularly relevant for PbRL, where both sample efficiency and robustness to noisy labels are key for practical deployment. In this experiment, we conduct a preliminary study to examine whether our framework can leverage VLM-based preferences effectively.
        The prompt and configuration provided to the VLM are shown in \figref{fig:gpt_prompt}. While VLMs offer a potential substitute for human preference annotations, they often tend to be overconfident when choosing between options and rarely select the ``Can't decide'' label. This behavior may stem from their underlying training or preference-oriented fine-tuning. To address this, as illustrated in \figref{fig:gpt_prompt}, we assumed that the VLM's frame-wise estimates are independent and identically distributed and applied Welch's t-test. If no statistically significant difference was found, the query was mapped to the ``Can't decide'' label. This experiment used the target feature value of ``Posture'' as defined in \tabref{table:feature_extractor}.

    \subsubsection{Results}
        As shown in \tabref{table:GPT}, VLM-based preference labels exhibit both discriminability and accuracy limitations. The Baseline achieved only about \SI{60}{\%} label accuracy, which is extremely low considering that the labels are binary. Moreover, nearly half of the queries could not be assigned a label, leading to almost no learning in terms of the feature error $d^{\text{pref}}$. These results indicate that VLMs tend to generate overconfident outputs, avoiding the ``Can't decide'' option and thus producing unreliable supervision.
        In contrast, our framework leverages preference discriminability to reduce the number of indistinguishable queries, improving label accuracy by approximately \SI{20}{\%} over the Baseline. Consequently, the learned policy reached the feature error threshold ($d^{\text{pref}} < 0.02$) defined in \figref{fig:ex1}, demonstrating that useful signals can be extracted from VLM-generated labels when query design accounts for discriminability.

\subsection{Scalability with Respect to Feature Dimensions}
\label{ex_task_scalability}
    \subsubsection{Settings}
        This section evaluates the applicability of the proposed framework to different tasks by testing its effectiveness on two additional task settings with a greater number of feature parameters. The experiment uses the target feature values of ``Trot Only'' and ``Normal'', as defined in \tabref{table:feature_extractor}. In these settings, the number of features required for learning from queries is increased to four and six, respectively, compared to two in the tasks used in the previous section, making the learning process more challenging. All experiments were conducted under the ``Medium'' preference discriminability condition.

    \subsubsection{Results}
\begin{figure}[t]
    \centering
    \includegraphics[width=0.92\columnwidth]{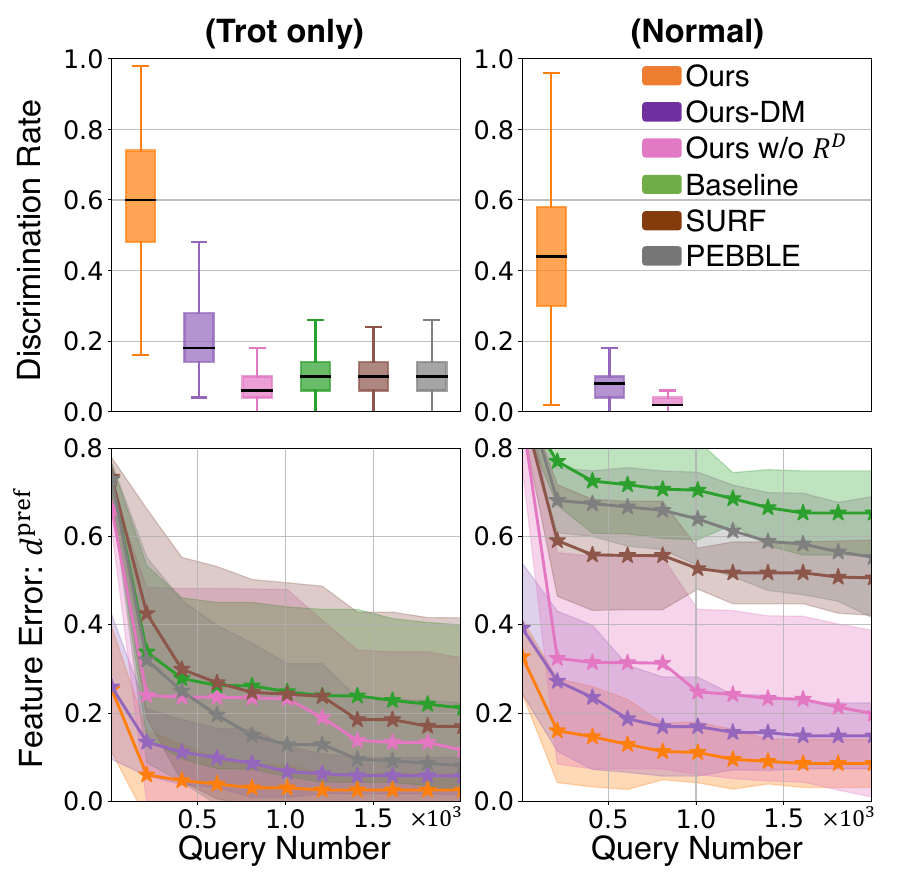}
    \vspace{-2mm}
    \caption{
        Comparison of query efficiency in various tasks:
        ``Trot only'' means using feature extraction of trot patterns, ``Normal'' adds feature extraction of body posture as shown in \tabref{table:feature_extractor}.
        This experiments use preference discriminability $d^{\text{disc}}$ = ``Medium'', meaning that if a policy's features are closer than a specified threshold, the policy is considered non-identifiable ($y=0.5$).
        Each curve plots the mean and standard deviation over five experiments. 
    }
    \label{fig:ex5}
\end{figure}

        As shown in \figref{fig:ex5}, Ours achieves faster convergence to the target feature $f^*$ across all tasks. 
        As the number of features $|f|$ increases, the difficulty of discrimination also rises, and the number of required queries tends to increase accordingly. In particular, previous methods struggle significantly with the ``Normal'' setting, which requires learning six features, resulting in very few identifiable queries. While the learning difficulty may ultimately depend on how close the policy needs to get to the target feature $f^*$, it is possible that even our proposed framework may face limitations as the feature dimensionality increases further. Investigating this 
        potential limitation remains an important direction for future work.

\begin{table}[t]
    \small
    \caption{
        Performance evaluation with VLM-based preference labeling:
        Each trial was conducted with up to 1,000 queries.
        Each value represents the mean and standard deviation over five experiments.
    }
    \label{table:GPT}
    \begin{center}
        \begin{tabularx}{\linewidth}{l *{2}{>{\centering\arraybackslash}X}}
            \toprule
            \textbf{Evaluation Metric} & \textbf{Ours} & \textbf{Baseline} \\ 
            \midrule
            Indiscriminable Rate [$\%$] & $23.6 \pm 15.7$ & $46.9 \pm 16.0$ \\
            Label Accuracy [$\%$]      & $84.7 \pm 12.8$ & $61.9 \pm 15.2$ \\
            Minimum Feature Error $d^{\text{pref}}$ & $0.005 \pm 0.002$ & $0.318 \pm 0.061$ \\
            \bottomrule
        \end{tabularx}
    \end{center}
\end{table}

\subsection{Real-World Application}
\label{ex_sim2real}

\begin{figure}[t]
    \centering
    \includegraphics[width=0.99\columnwidth]{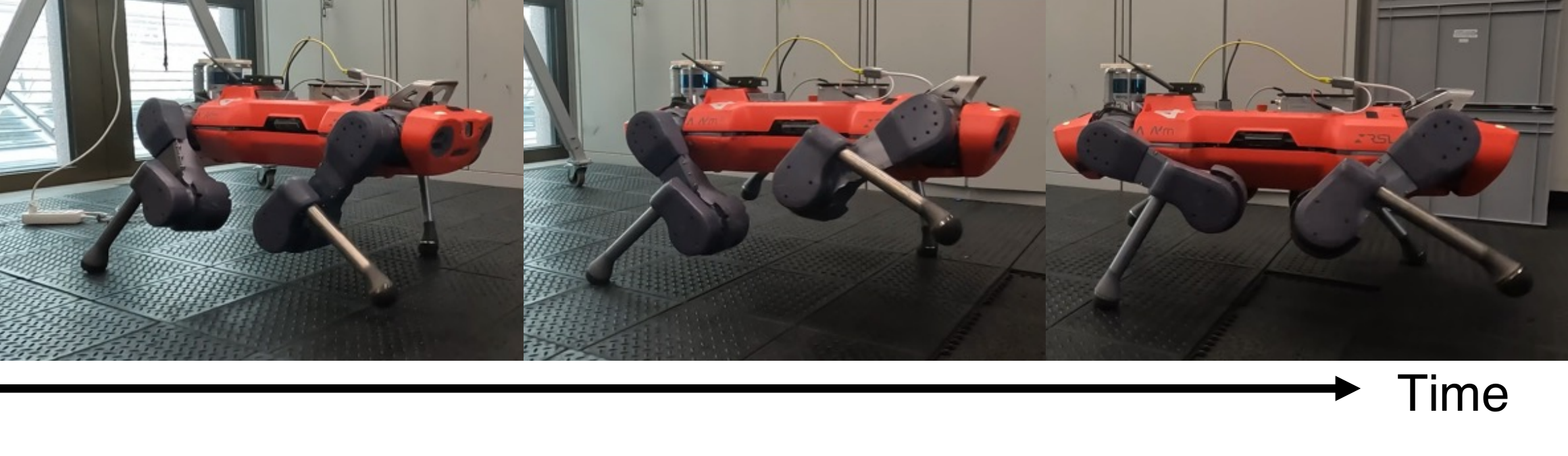}
    \vspace{-7mm}
    \caption{
        Real-world execution of policies trained in simulation:
        The ``Crawl'' walking pattern listed in \tabref{table:feature_extractor} is trained in simulation, and the learned policies were deployed on the ANYmal quadrupedal robot.
    }
    \label{fig:ex6}
\end{figure}

    In this section, we apply the learned control policies from simulation to real-world environments using a sim-to-real transfer approach \cite{legged_gym}.
    The transfer policy is trained in a domain-randomized environment over sensor noise and physical parameters, using well-validated parameter sets from previous work \cite{legged_gym}.
    During policy learning, the feature representation $f$ used as input to the learned preference reward model $R^H$ is treated as privileged information available only in simulation for reward computation, and therefore domain randomization is not applied to $f$.
    As shown in \figref{fig:ex6}, the learned policies were applied in the real world.
    This validation follows the same sim-to-real setup as in previous work \cite{legged_gym}, without additional tunings.
    For details of the robot motion, please refer to the supplementary videos.

\section{Discussion}
\label{s:dis}

    \textbf{Robustness to Label Noise in Preference Annotations:}
        This paper focuses on preference discriminability and proposes a robust learning framework that performs effectively even under high discriminability thresholds (i.e., ``Large''). In practice, however, some queries within these thresholds may not be labeled as indistinguishable by human annotators but instead receive incorrect superiority labels, as shown in \figref{fig:ex4}(c).
        In addition, as shown in \tabref{table:GPT}, we observed that the proposed framework can still operate effectively even when the generated labels include a certain amount of mislabeling (approximately \SI{15}{\%}), suggesting robustness to imperfect annotations.
        This type of labeling noise has also been studied extensively in several prior works \cite{pbrl_noisyPreference, pbrl_misalignment}. Building on these insights, integrating such noise-robust methods with our framework presents a promising direction for future research.

    \textbf{Sensitivity to Hand-crafted Feature Design and Future Direction Toward Automated Extraction:}
        Some consideration is necessary when defining the terms of feature extraction \cite{feature_extraction_pbrl}. In particular, the success of our framework still relies on a set of carefully designed features, and the overall performance is sensitive to the quality of this feature selection. If multiple similar features are selected and their optimization objectives happen to conflict, it may hinder policy learning. For example, if satisfying different features requires the robot to exhibit opposing behaviors, achieving the target may become difficult or even infeasible. Likewise, if the selected features are noisy or fail to capture informative aspects of user preferences, the learned policy can become suboptimal. While such cases are unlikely in well-structured tasks, they remain a possibility, and attention to feature design is crucial. A promising future direction is to reduce this dependence on hand-crafted features by integrating automated feature extraction or richer representations learned from high-dimensional sensor data.

    \textbf{Trade-off Between Computational Cost and Policy Diversity:}
        An additional aspect to consider is the computational cost associated with retraining a new policy from scratch after each reward update in order to learn policies with different feature representations. 
        While this approach is more expensive than continuing from previously trained policies, it was adopted to avoid policy bias and to ensure diversity among the learned policies. 
        This highlights a trade-off between computational efficiency and policy diversity: retraining increases training time but provides more reliable exploration and reduces the risk of converging to biased behaviors. 
        We recognize that this trade-off becomes a critical concern when extending the framework to more complex and computationally demanding domains, such as large-scale decision-making problems that require semantic or linguistic understanding.
        To mitigate this limitation, potential directions include warm-starting policy initialization within a range that avoids policy bias, or leveraging meta-learning techniques to initialize policies in a cost-effective manner.
        These strategies could help reduce computational overhead while maintaining sufficient diversity across learned policies, making the framework more scalable to large-scale or high-dimensional applications.

    \textbf{Balancing Low-level Constraints and High-level Preferences:}
        In our framework, policy constraints (\tabref{table:policy_constraint}) are used to enforce basic requirements such as movement speed or safety, rather than to represent human preferences. These constraints do not need to be strictly maximized, but only sufficiently satisfied within a margin. This approach is particularly effective because such low-level objectives are easy to define and help improve query discriminability. In contrast, high-level preferences such as motion style are difficult to specify explicitly, and are therefore learned through PbRL. This separation allows the system to delegate well-defined tasks to constraints, while using PbRL to capture complex, preference-based behaviors.

    \textbf{Stability of the Discriminability Reward During Early Training:}
        A potential concern in our framework is the stability of the discriminability reward $R^D$ during the early stages of training. Since the discriminator is initially trained on only a small number of labels, its preference discriminability estimates may be noisy or biased, which could misguide policy exploration and lead to suboptimal behaviors. This sensitivity can become more noticeable as the input and output dimensionality of the discriminator increases, making it harder to obtain reliable preference discriminability scores from limited data. In our experiments, however, such instability was not severe, partly because the feature dimensionality was relatively small and the discriminator $D$ incorporated Monte Carlo dropout \cite{mc_dropout}, which helped mitigate instability in its outputs. Nevertheless, instability remains a possible challenge in more complex scenarios, and developing mechanisms to further stabilize the $R^D$ represents an important direction for future work, for example by incorporating dimensionality reduction techniques or attention mechanisms.

    \textbf{Toward Capturing Richer and Higher-dimensional Human Preferences:}
        While our framework takes a step toward making PbRL more practical by explicitly addressing query discriminability, it still falls short of fully capturing human preferences. The feature dimensionality in our experiments (up to six) may overlook subtle aspects of human preference, such as smoothness, energy efficiency, or style. Moreover, learning a satisfactory policy still requires hundreds of queries, although \figref{fig:ex1} shows that our framework reduces this number to about one-tenth of that required by previous approaches. These findings indicate that further progress toward real human experiments will require richer, higher-dimensional preference features and additional techniques to reduce the query burden.
        In addition, when features become higher-dimensional, humans may perceive a wider threshold of indistinguishability, which corresponds to a larger threshold of $d^{\text{disc}}$.
        This phenomenon was also observed in \figref{fig:threshold}, where increasing the threshold step by step beyond the real human range (\figref{fig:ex4}) eventually caused the our framework to fail to learn policies.
        Developing mechanisms that can adaptively handle such broader indistinguishability ranges will be an important direction for future research.

    \textbf{Limitations and Opportunities of VLM-based Preference Annotation:}
        Another important point emerging from our VLM-based experiments is the challenge of relying on VLMs as annotators. Limitations such as weak spatial understanding and overconfidence in discrete choices restrict their current reliability for preference labeling \cite{vlm_spatial,vlm_asReward}. Our evaluation should therefore be regarded as a basic proof-of-concept, representing an initial step to verify feasibility. While preliminary, these findings highlight both the potential and the current limitations of integrating VLM feedback, and they point to opportunities for future research on more robust methods in PbRL.

    \textbf{Domain Scope and Future Extensions Beyond Quadruped Locomotion:}
        Our experiments have primarily focused on quadruped locomotion tasks, as this domain provides a representative and challenging benchmark for evaluating PbRL frameworks. Quadruped robots require learning diverse gait patterns under high-dimensional dynamics, making them a suitable testbed for studying query efficiency and policy diversity. While the proposed framework is not limited to locomotion and could, in principle, be extended to other robotic domains such as manipulation or humanoid control, a thorough exploration of these directions lies beyond the scope of this paper. We consider this an important avenue for future work, as demonstrating the applicability of our framework across a broader range of skills and platforms would further strengthen its generality.

\section{Conclusion}
\label{s:con}
    This paper proposed DAPPER, a framework that improves query efficiency by incorporating human preference discriminability and trajectory diversification achieved by multiple policies.
    DAPPER trains multiple policies and learns a discriminator to generate queries that are easier for humans to label, addressing the limited diversity problem caused by policy bias inherent in previous trajectory-based approaches.
    Experiments on simulated and real-world legged robot tasks demonstrated that DAPPER significantly improves query efficiency compared to previous methods, particularly under challenging discriminability conditions.

\addtolength{\textheight}{-0cm}   


\bibliographystyle{IEEEtran}
\bibliography{IEEEabrv,reference}

\end{document}